# Recognizing Activities and Spatial Context Using Wearable Sensors

**Amarnag Subramanya**
Electrical Engineering
University of Washington
Seattle, WA 98195

**Alvin Raj**
Computer Science & Engineering
University of Washington
Seattle, WA 98195

**Jeff Bilmes**
Electrical Engineering
University of Washington
Seattle, WA 98195

**Dieter Fox**
Computer Science & Engineering
University of Washington
Seattle, WA 98195

## Abstract

We introduce a new dynamic model with the capability of recognizing both activities that an individual is performing as well as where that individual is located. Our approach is novel in that it utilizes a dynamic graphical model to jointly estimate both activity and spatial context over time based on the simultaneous use of asynchronous observations consisting of GPS measurements, and a small mountable sensor board. Joint inference is quite desirable as it has the ability to improve accuracy of the model and consistency of the location and activity estimates. The parameters of our model are trained on partially labeled data. We apply virtual evidence to improve data annotation, giving the user high flexibility when labeling training data. We present results indicating the performance gains achieved by virtual evidence for data annotation and the joint inference performed by our system.

## 1 Introduction

Recent advances in wearable sensing and computing devices and in fast probabilistic inference techniques make possible the fine-grained estimation of a person's activities over extended periods of time [12]. Such technologies enable applications ranging from context aware computing [9] to support for cognitively impaired people [16] to long-term health and fitness monitoring to automatic after action reporting of military missions.

The focus of our work is on providing accurate information about a person's activities and environmental context in everyday environments based on wearable sensors and GPS devices. More specifically, we wish to estimate a person's motion type (such as walking, running, going upstairs/downstairs, or driving a vehicle) and whether a person is outdoors, inside a building, or in a vehicle. These activity estimates are additionally combined with GPS information so as to estimate the trajectory of the person along with information about which buildings the person enters. To do this, our approach assumes that the bounding boxes of buildings (i.e., maps) are known. This assumption is reasonable, given the availability of satellite images from which these bounding boxes can be extracted manually or via computational vision algorithms.

Our main emphasis is on performing activity recognition that is not only accurate, but that also requires a minimum number of sensor devices. There are in fact a variety of systems that utilize multiple sensors and measurements taken all over the body [11, 15]. Our approach, by contrast, attempts to produce as accurate as possible activity recognition requiring only one sensing device mounted only at one location on the body. Our reasoning for reducing the total number of sensors is threefold: 1) it can be unwieldy for the person wearing the sensors to have many such sensors and battery packs mounted all over the body, 2) we wish to minimize overall system cost, and 3) we wish to extend operational time between battery replacement/recharge (equivalently, minimize online energy consumption).

Our model is novel in several respects. We integrate sensor information from both a standard GPS device and a single sensor board (producing standard signals such as accelerometer, barometric pressure, etc.) in a semi-asynchronous way. In other words, it is not required for our sensor board to produce measurements at precisely the same time points as does the GPS device. This approach therefore allows us to utilize separate, standard, cost-effective, and non-integrated components for multiple types of information, thereby reducing the overall cost of the system. Our statistical model, moreover, utilizes a dynamic Bayesian network (DBN) [7] for expressing the underlying dependencies between various types of information represented in our system, and performs efficient probabilistic inference based on a discretization of 2-D space. State space pruning allows us to perform joint inference on both a person's location and her activity.

In Section 2, we give a brief overview of our sensor board.

Section 3 describes our activity model including all modeling assumptions, inference, and learning algorithms. Experiments are described in Section 5, followed by a discussion and conclusions.

## 2 Wearable Sensor System

Our customized wearable sensor system consists of a multi-sensor board, a Holux GPS unit with SIRF-III chipset, and an iPAQ PDA for data storage.

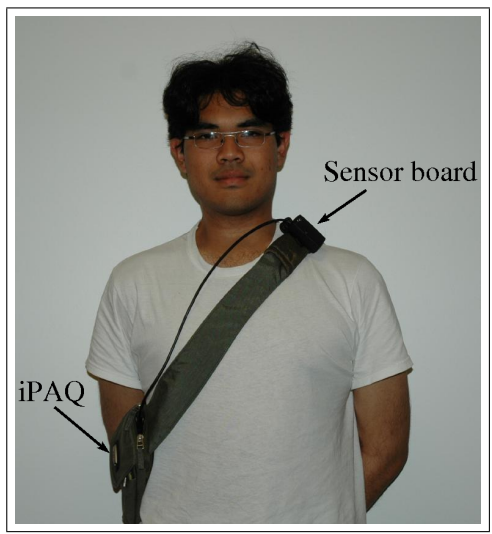

Figure 1: Multi-sensor board and PDA used in our system.

The sensor board shown in Figure 1 is extremely compact, low-cost, and uses standard electronic components. It weighs only 121g *including battery and processing hardware*. Sensors include a 3-axis accelerometer, two microphones for recording speech and ambient sound, phototransistors for measuring light conditions, and temperature and barometric pressure sensors. The overall cost per sensor board is approximately USD 400. The time-stamped data collected on this device is transfered via a USB connection to an iPAQ handheld computer. GPS data is transfered from the receiver via Bluetooth to the PDA. The overall system is able to operate for more than 8 hours.

## 3 Activity Model

### 3.1 Overview

The complete DBN for our activity model is shown in Figure 2 with some of the implementation details shown in Figure 3. The variables represented in the model include GPS measurements $(g_k, h_k)$, sensor-board measurements $m_k$, the person's location $l_k$, motion velocity $v_k$, the type of motion $s_k$ they are performing, and their environment $e_k$. We first describe the individual components and their relationships starting at the sensor level of the model.

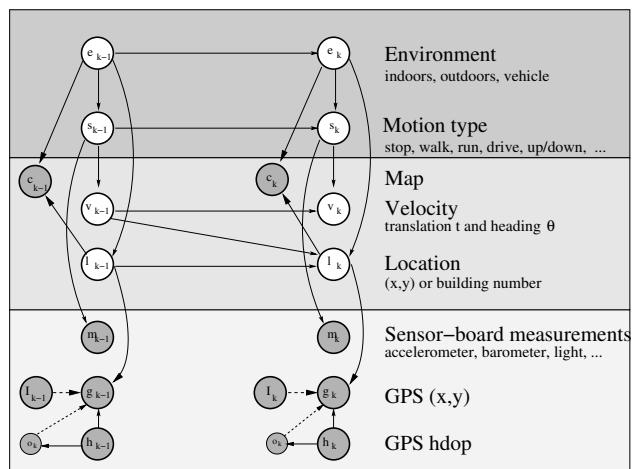

Figure 2: Graphical model representation of the activity recognizer.

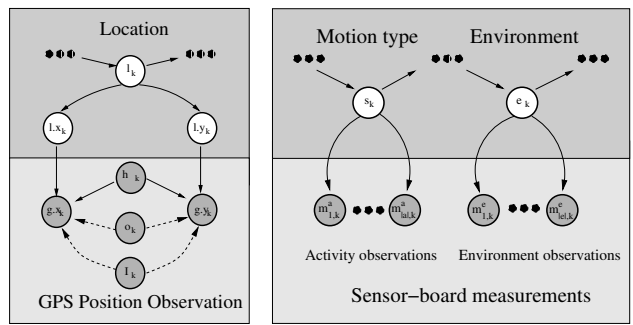

Figure 3: Relationships (left) between locations and GPS measurements, and (right) between state, environment, and sensor-board measurements.

**GPS measurements** are separated into longitude / latitude information, $g_k = \{g.x_k, g.y_k\}$, and *horizontal dilution of precision* (hdop), $h_k$. hdop provides information about the accuracy of the location information, which depends mostly on the visibility and position of satellites. The node $o_k$ explicitly models GPS outliers. Outliers typically occur when the person is inside or close to a building or under trees. Unfortunately, outliers are not always indicated by a high hdop value, especially when the person is very close to high rise buildings.

The left panel in Figure 3 shows a more 'low-level' view of the sub-graphical model for the GPS observations. In this figure $g.x_k$ and $g.y_k$ are the GPS northing and easting readings, respectively. These are scored by the northing and easting of the current location $l_k$, namely $l.x_k$ and $l.y_k$, and the current value of hdop $h_k$. It should be noted that both $l.x_k$ and $l.y_k$ are deterministic children of the hidden variable modeling location, i.e., $l_k$. Also, $o_k$ is an observed switching parent such that when $o_k = \{1\}$, the graph is exactly as shown in Figure 3, but when $o_k = \{0\}$, the GPS observations are disconnected from all the their (non-switching) parents. Furthermore, the two observation streams in our system, namely, the sensor board and the GPS, are not synchronized. In particular, the GPS is the less frequent of the two. In order to handle this, we make use of a switching parent $I_k$, which disconnects $g_k$ from all its parents when $I_k = 0$. The likelihood of a GPS measurement is given by

$$p(g_k|l_k, h_k, o_k) = \mathcal{N}(g_k; l_k, \sigma_{h_k}^2) \text{ if } o_k = 1 \text{ and } I_k = 1 \quad (1)$$

That is, if the GPS observation is present at time $k$, and is not an outlier, then the likelihood is given by a Gaussian centered at the person's location. The variance of the Gaussian is a function of the hdop value $h_k$. In our current implementation, we have found that setting $\sigma_{h_k}^2 = 2 * h_k$ works reasonably well.

**Sensor-board measurements** $m_k$ consist of pertinent features extracted from 3D acceleration, barometric pressure, temperature, visible and IR light intensity, and raw audio. As we will further describe in Section 3.4, we use boosted classifiers to extract probability estimates for the person's instantaneous environment and motion state. These classifier outputs provide the observation $m_k$, which in our model depends on the current environment and motion state, as indicated by the arcs from $e_k$ and $s_k$. More specifically, we learn one binary classifier for each motion state and environment. At each iteration, each of these classifiers generates a probability estimate $p_i$ that the current measurement corresponds to the state or environment it was trained on (see right panel in Figure 3). We then uniformly quantize the $p_i$'s into $b$ bins. This gives us an integer observation vector $m_k = (m_k^s, m_k^e)$ at time $k$, where $m_k^s = (m_{1,k}^s, \ldots, m_{|s|,k}^s)$, and $m_k^e = (m_{1,k}^e, \ldots, m_{|e|,k}^e)$ where, every $\alpha \in m_k$ is such that $1 \leq \alpha \leq b$, $|s|$ and $|e|$ are the cardinalities of the state (activity) and environment variables respectively. In our current implementation, $b = 10$, i.e. each $p_i$ is uniformly quantized into 10 bins.

We assume each discretized binary classifier observation is independent of all else given the corresponding environment or motion state, and along with a naive Bayes like assumption, obtain the following observation model:

$$\begin{aligned} p(m_k \mid s_k, e_k) &= p(m_k^s \mid s_k) \, p(m_k^e \mid e_k) \\ &= \prod_{i=1}^{|s|} p(m_{i,k}^s \mid s_k) \prod_{i=1}^{|e|} p(m_{i,k}^e \mid e_k) \quad (2) \end{aligned}$$

Note that in our current implementation, we assume that the observations from the sensor board are available all the time.

**Map** There are many ways of expressing constraints in Bayesian networks [8], each of which can be implemented using a variety of algorithms. We choose in this work to utilize the mechanism of the observed child. Specifically, we encode position and map information using an observed child $c_k$ that implements a constraint between its two parents, namely, the current location $l_k$ and the environment $e_k$. This variable and its conditional probability table (CPT) represents knowledge about the locations of buildings in the area under consideration. Values in the CPT can be used to enforce consistency between the assignments to the location and environment variable in either a soft or hard manner. For example, we can easily enforce the constraint that when $l_k$ is outside any of the buildings in the map, a value of $e_k = inside$ would yield zero probability, in the form of $p(c_k = 1|e_k = inside, l_k) = 0$ for appropriate values of $l_k$, and where the map is encoded in the CPT $p(c_k|e_k, l_k)$. The above is an example of a hard constraint, but it is straightforward to extend the above to soft constraints in which case sets of variable assignments would get probabilities in this CPT other than 0 or 1.

**Location** In reality the location of a person is a continuous function of time. In [18], we show how to apply Rao-Blackwellized particle filters in order to perform approximate inference using a sample-based representation of the state space. In this paper, however, we simplify the problem by discretizing the location variable. We assume that we have a fixed region over which the user moves, which is then discretized using fixed square grids of size $100cm$. Currently, we do not aim to track people inside buildings since GPS is either useless or not available, and because the accelerometers on our sensor boards are not sufficiently accurate to provide adequate dead reckoning information. The location at time $k$ solely depends on the person's previous location $l_{k-1}$ and motion $v_{k-1}$. Note that location is hidden during both training and testing.

**Velocity** represents the motion between locations at consecutive points in time. We adopt a piecewise linear motion model on polar velocity coordinates $v_k = (t_k, \theta_k)^T$, namely translational speed $t_k$ and heading direction $\theta_k$. Once again we discretized the number of possible speed and heading values. We assume that the heading at time $k$ only depends on the previous heading and the current activity. By making the change in heading also depend on the activity, our system can model the fact that, for example, a walking user changes heading differently than a user driving a vehicle. The translational speed $t_k$ is modeled in a similar fashion to the heading, and depends on the previous speed $t_{k-1}$ and the current motion state $s_k$. In order to aid data sharing across activities, both the CPTs, i.e., speed and heading are implemented using sparse representations. This is because the motion patterns for various classes of activities are similar. For example, walking and going up/downstairs are similar, as going up/downstairs are essentially ramifications of walking on inclined surfaces. Sharing motion patterns across the above set of activities not only aids better training, but also speeds up inference.

**Motion states** represent different types of motion a person can be involved in. In our current system, these states include $s = \{$`stationary, walking, running, driving vehicle, going up/down stairs`$\}$. The motion state $s_k$ depends on the previous motion state $s_{k-1}$ and the current environment $e_k$. The temporal dependency allows the system to capture information such as "it is very unlikely to get into the driving state right after going upstairs". It should be noted that this variable is observed during training, as we assume that we have frame level labels [1].

**Environment** captures the person's spatial context, which is $E = $ indoors, outdoors, vehicle. Note that due to the edge between $e_k$ and $s_k$, there can be (both soft and hard) constraints imposed between the motion state and the environment. For example, whenever it is the case that the environment is in the indoors or outdoors state, we *a priori* preclude driving from being a possible value of the motion type (i.e., it has zero probability). Whenever the environment is in the vehicle state, the motion type may not be up/down stairs (but it may be stationary, for example). Moreover, other "soft constraints" are imposed by the fact that the two nodes are related probabilistically, and the probabilities are learned automatically (see Section 3.2). Like the motion state variable, the environment variable is observed during training.

### 3.2 Inference

Our DBN-based probabilistic inference system uses a combination of a triangulation [4] (to produce a dynamic junction tree) and search-based methods within each clique. This allows us to utilize a form of approximate inference method that is akin to sampling, but has some differences. As each clique is constructed, we only use a subset of the states of the incoming separators. The subset is chosen indirectly using a pruning strategy — specifically, for each separator, its previous clique has had all but the top scoring $k$ clique-states removed. This leads to a reduced separator size and thus a reduced within-clique construction procedure.

As a result, our inference system adaptively prunes the state space. Since the pruning threshold is chosen based on the probability of the most likely state, the number of states being tracked depends on the uncertainty of the posterior; the higher the uncertainty, the less states are pruned. This pruning strategy is essential for tractable inference and learning, since the overall state space of our model contains on an average over $100,000$ states for each time frame. If left unpruned, the state space can potentially increase exponentially over time. In our experiments, the system typically tracks only about 10,000 states, without noticeable reduction in inference and learning accuracy. Further details of this DBN inference procedure will appear in forthcoming publications. All DBN inference and incorporation of dynamic virtual evidence (see below) utilizes the *graphical modeling toolkit* (GMTK) system [4].

---

[1] Note that in section 3.3 we discuss the case when frame level labels are not available

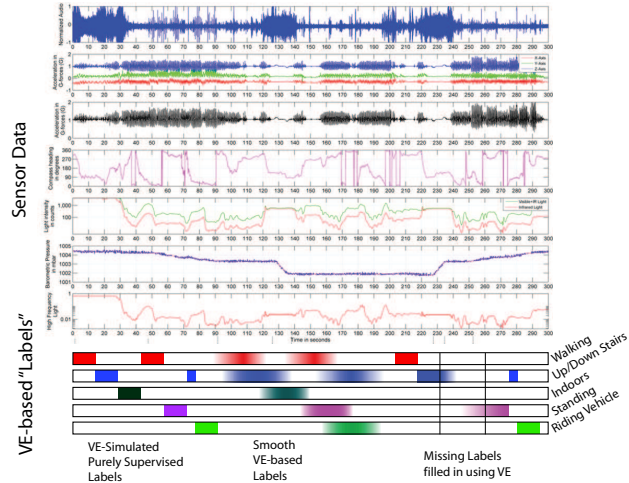

Figure 4: Example annotations enabled by virtual evidence (VE). The system can model arbitrary levels of uncertainty in current states and transition times.

### 3.3 Semi-supervised Learning With Virtual Evidence

The robustness and accuracy of activity recognition models largely depends on the availability of labeled training data. A typical approach to learning such models is to collect long sequences of training data and to manually label them with the performed activities. Unfortunately, such an approach is extremely tedious and error-prone. Annotation errors are typically due to confusing different activities and due to wrong timing of transitions between activities. The second type of error is especially difficult to overcome since it is often unclear or not well defined when a transition actually occurs, or what a transition means. For instance, the exact time point when a person enters the driving vehicle state can be chosen in various ways, for instance when the person enters the car, when she sits in the car, when she turns the motor on, or when the car starts to move.

An obvious solution to this problem would be unsupervised training of the model parameters. Unfortunately, such an approach typically generates models that do not directly correspond to the set of activities one wants to detect. A popular solution is to use a mixture of supervised and unsupervised (i.e., semi-supervised) learning [19, 5, 6], where parts of the data are labeled and other parts are left unlabeled. While such an approach provides reasonable results, it gives the user only a limited set of options when labeling data (fully, or not at all).

In our system, we use the notion of virtual evidence (VE) to model a variety of data annotations. Virtual evidence was first defined in [17] as a way to depict evidence in a Bayesian network using the mechanism of hidden variables. Given a joint distribution over $n$ variables $p(x_1, \ldots, x_n)$, evidence simply means that one of the variables (w.l.o.g. $x_1$) is observed. We denote this by $\bar{x}_1$, so the probability distribution becomes $p(\bar{x}_1, \ldots, x_n)$ which is no longer a function of $x_1$ since it is fixed. Essentially, any configuration of the variables where $x_1 \neq \bar{x}_1$ is never considered. We can mimic this behavior by introducing a new virtual child variable $c$ into the joint distribution that is always observed to be one (so $c = 1$), and have $c$ interact only with $x_1$ via the CPT $p(c = 1|x_1) = \delta_{x_1 = \bar{x}_1}$. Therefore, $\sum_{x_1} p(c = 1, x_1, \ldots, x_n) = p(\bar{x}_1, \ldots, x_n)$.

More general virtual evidence is formed where $c$ interacts with two or more variables, and rather than using a delta function, we use a "soft evidence" function. That is, suppose $c$ interacts with $x_1$ and $x_2$. We set $p(c = 1|x_1, x_2) = f(x_1, x_2)$, where $f()$ is an arbitrary non-negative function.[2] Therefore, additional preferential treatment can be given to different combinations of $x_1$ and $x_2$, but unlike hard evidence, we are not insisting on only one particular pair of values. Virtual evidence can be used to incorporate soft evidence when computing posterior distributions over the most probable values of random variables (filtering, Viterbi, or MPE), and during EM training [3].

In this work, we employ virtual evidence as a mechanism for semi-supervised learning. Clearly, making $f()$ a delta function mimics evidence. And making $f()$ the same score for all values of its arguments mimics an entirely hidden variable. These two extremes capture the typical case in semi-supervised learning. Under neither of these extremes, however, $f()$ can provide numerical preference for or against some set of values of the hidden variables reflecting a graded uncertainty about a particular label.

We can use the VE mechanism to represent uncertainty in the supervisory training procedure in the ways outlined at the beginning of this section. Figure 4 illustrates some of the annotations allowed in our system. First, on the left of the figure shows standard supervised labels being simulated by VE, where at any given time, $f()$ provides a non-zero score only to one possible activity. Second, we may wish to model the fact that the labels at transitions are typically unreliable. We thus can add a smoothly varying time-inhomogeneous collection of virtual evidence functions, moving from one class label to the next — these are shown as the center set of labels in the figure. For example, if we decide that a transition starts at time $k_1$, ends at time $k_2$, and makes a transition from label $\ell_1$ to $\ell_2$, we utilize a time-dependent function $p(c_k = 1|a_k) = f_k(a_k)$ where $a_k$ is the activity at time $k$, and where $f_k()$ varies smoothly between the extremes of insisting on label $\ell_1$ to the same on label $\ell_2$. By putting less stringent requirements on the posteriors, training can result in parameters that have a reduced variance. Third, we may have regions that are entirely missing labels, as shown as the right set of labels in the figure. In this case, we can again utilize an $f()$ function that gradually "fades out" from the most recent label, and fades in to the next label. There are a number of ways to do this. For example, $f()$ can fade out to the point where it provides uniform score over all activities. Alternatively, it could fade out to the point where it only provides uniform score to the previous and the following known activity. We experiment with these approaches in Section 5.

### 3.4 Feature Extraction

Our system uses GPS measurements and sensor-board measurements to infer a person's activities. Especially the sensor board produces a variety of discrete signals that have different properties, such as inherently different sampling rates and spectral contents — for example, raw audio has a much higher sampling rate than the barometric pressure sensor. To incorporate such heterogeneous sensor data, we employ the feature extraction process developed by Lester and colleagues [12]. Here, we only provide a high-level description of this process, more information can be found in [12].

First, the signal sample rates are normalized by low-pass filtering and/or up/down-sampling to an appropriate rate so that information is not lost. Next, each signal is windowed using a window of appropriate length at 4 Hz, and in each window a feature vector is extracted, resulting in a feature vector of high dimensionality (650 in our current system). Such high-dimensional feature vectors are not possible to utilize directly in a model, and typical approaches either require dimensionality reducing linear transforms such as PCA or LDA, or alternatively feature selection. We utilize the approach taken in [12] as our starting feature extraction procedure. Essentially, for each activity we learn boosted threshold detectors, or decision stumps. In other words, for each activity, we learn a collection of decision stumps where the next stump is obtained via boosting. Each decision stump essentially acts as a weak-learner, alone incapable of making an accurate detection decision, but when combined with kindred classifiers, capable of making highly accurate decisions. This collection of decision stumps for each activity is then used to produce a final event detection probability $0 \leq p_i \leq 1$ for activity $i$ (as mentioned in Section 3.1). The detection probability $p_i$ is obtained by viewing each individual threshold as a decision boundary, the distance to which constitutes a margin. Considering these margins together, we can obtain an average distance to the decision boundary, which is then passed through a sigmoid function to produce a $[0, 1]$-valued probability. It is these probabilities that are then uniformly quantized to produce the integer observations — e.g., $m_{i,k}^a = 3$ if $0.2 \leq p_i < 0.3$. In future work, we plan to utilize standard joint vector-quantization algorithms for this last process, but for now we found that the above simple approach worked well for our applications.

---
[2] Normalization is performed whenever a state distribution needs to be extracted from the system.

## 4 Related Work

Recently, the estimation of activities from wearable sensors has received significant attention especially in the ubiquitous computing and artificial intelligence communities. Virtually all existing approaches incorporate either location sensors such as GPS or location-independent sensors such as accelerometers and microphones, but not both.

Bao and Intille [2] use multiple accelerometers placed on a person's body to estimate activities such as standing, walking, or running. Kern and colleagues [11, 10] and Lukowicz *et al.* [15] added a microphone to a similar set of accelerometers in order to extract additional context information. These techniques typically rely on Gaussian observation models and dimensionality reduction techniques such as PCA and LDA to generate observation models from the low-level sensor data or features extracted thereof (*e.g.*, FFT). Recently, Lester *et al.* [12] showed how to apply boosting in order to learn activity classifiers based on the sensor data collected by the same sensor board we use in our research. These existing approaches feed the sensor data or features into static classifiers [2, 9], a bank of temporally independent HMMs [12], or multi-state HMMs [10, 11] in order to perform temporal smoothing. None of these approaches estimates a user's spatial context or is able to consider such context so as to improve activity recognition.

Using location for activity recognition has been the focus of other work. For instance, Ashbrook and Starner [1] detect a person's significant places based on the time spent at a location. In [13], Liao and colleagues showed how to learn a person's outdoor transportation routines from GPS data. More recently, the same authors presented a technique for jointly determining a person's activities and her significant places [14]. However, all these approaches are limited in their accuracy due to the fact that they only rely on location information.

In contrast to existing work, our system combines the information from both classes of sensors. We apply a structured DBN model in order to efficiently perform joint inference over our more complex state space. The ability to incorporate multiple modalities into the estimation process results in a system that is able to extract more expressive context information from the sensor data.

## 5 Experiments

To evaluate our system, multiple users were asked to perform a variety of activities on the University of Washington campus. These activities included walking, running, going up/down a flight of stairs, driving around in a vehicle, going indoors, etc. Users were instructed to perform the above activities in a natural way and neither the sequence of activities nor their durations was choreographed. The users were asked to label the activities they performed using a hand-held device, *i.e.*, each time a user started a new activity, she had to indicate that activity on the PDA.

In our current implementation we make use of a rather simple GPS outlier detector. We simply ignore GPS whenever the hdop value $h_k > 8$. As we have a mixture of both continuous and discrete observation distributions in this model, the scores from the continuous GPS distributions tend to outweigh the scores from the discrete observation distributions. Thus we raise the scores of the GPS observation distribution to a heuristically chosen exponent $0.5$ in the probability domain.

We extracted the locations of buildings from satellite images of the UW campus. A soft implementation is more prudent in this scenario because the reliability of the location estimate is very low in the absence of GPS (for example when the user is inside the building) and thus a hard map constraint can lead to undesirable results. In our current implementation, $p(c_k = 1 \mid e_k = inside, m(l_k) = inside) = 0.6$ and $p(c_k = 1 \mid e_k = outside, m(l_k) = outside) = 0.85$, where $m()$ is a (deterministic) function that maps a given location to either inside or outside a building based on the map information. These values were obtained after doing a grid search over the likely values that maximize the accuracy over a development set that was independent of the training and test sets. Note that the above soft map implementation may also be seen as using the map information as virtual evidence on the environment and location variables. The above model was trained using the EM algorithm until convergence.

In all there were 6 participants in the data collection effort, resulting in 25 data traces. Each trace had an average duration of 25 minutes. We performed leave-one-out cross-validation on these data sets. In each experiment, we trained the binary adaboost classifiers and discretized the margins of the weak learners, as explained in section 3.4. These discrete features and the GPS data were then used to jointly learn the parameters of the graphical model. The models were then evaluated based on the Viterbi output on the test trace.

Figure 5 shows a typical trace. The bounded regions in the figure represent buildings. The GPS observations are shown as circles connected by a black line. The most likely path as estimated by the system is shown using the thick, grey line when the environment is outside, and by stars when the environment is inside. It can be seen that the model is able to successfully trace the GPS and determine when and where the person enters and exits the building. Furthermore, it can be seen that the system successfully corrects the erroneous GPS measurements occurring near one building (such outliers are often not indicated by a high hdop value).

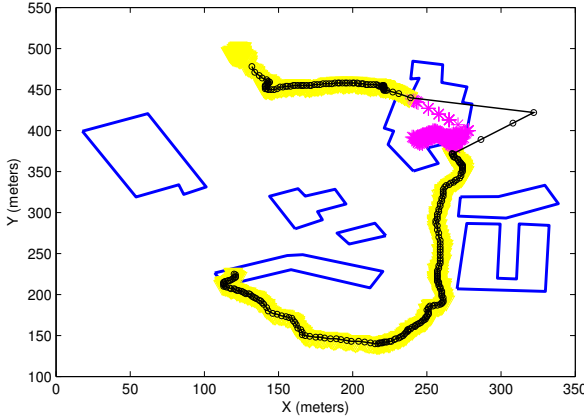

Figure 5: Example trace of the Viterbi output along with GPS locations and bounding boxes of buildings.

## Accuracy

For each trace, accuracy was determined by counting the number of correctly labeled frames divided by the total number of frames. We separately determined accuracy in estimating the person's motion state and accuracy in estimating the environment. The mean and 95% confidence intervals of the motion state and environment accuracies achieved on the 25 test traces are summarized in Table 1 and 2.

| Technique | Accuracy [%] |
|---|---|
| AdaBoost (MSB) | 77.0 ±2.5 |
| State only (MSB) | 82.1 ±2.1 |
| State + Environment (MSB) | 82.2 ±2.1 |
| State + Environment (MSB, GPS) | 86.8 ±1.9 |
| State + Environment (MSB, GPS, Map) | 86.9 ±2.0 |

Table 1: Accuracy in determining motion states.

| Technique | Accuracy [%] |
|---|---|
| AdaBoost | 83.8 ±3.7 |
| Environment only (MSB) | 88.7 ±3.7 |
| State + Environment (MSB) | 89.4 ±3.3 |
| State + Environment (MSB, GPS) | 91.5 ±2.8 |
| State + Environment (MSB, GPS, Map) | 93.02 ±2.4 |

Table 2: Accuracy in determining environment.

Each row in the tables provides the accuracy for a different model. AdaBoost is the output of the boosted classifiers on the data collected by the multi-sensor board (MSB). The second row in each of the tables shows the accuracies when a HMM was used to infer the state or environment independently. Note that this system only incorporates the boosted classifiers as observations, no GPS information is used. This system is similar to the technique used to infer activities in [12]. As can be seen, the temporal transition models provided by the HMMs result in about 5% performance gain in both state and environment tasks. Slight improvements can be gained by performing joint inference over the motion state and environment, as given in the third row of each of the tables.

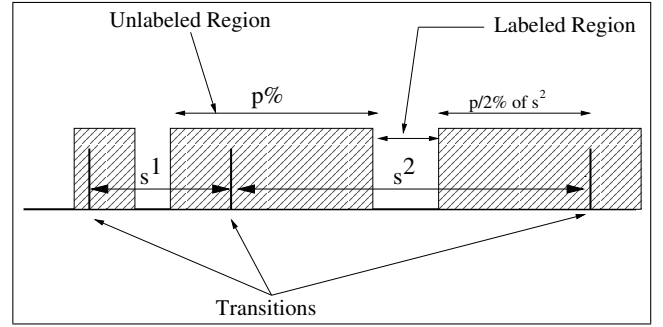

Figure 6: Illustration showing how we drop $p\%$ of the labels in a given trace.

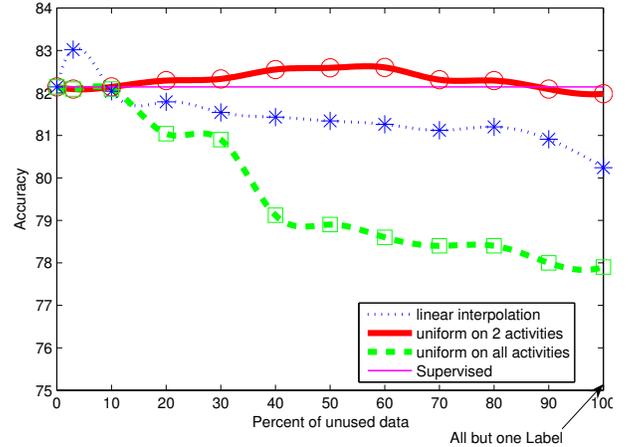

Figure 7: Results for using virtual evidence for partially labeled data.

The next row shows the result of fusing the MSB and GPS sensor streams. As can be seen, making use of GPS information in conjunction with the sensor board performs better than using the sensor board alone (more than 4% for motion state). This increase is mostly due to the fact that the GPS sensor provides additional information about a person's motion velocity. Finally, we present the results of adding information about the locations of buildings to the system. Although the results indicate that adding this information does not yield a significant improvement in state accuracy, we found that this system outperformed the system without map information in about 86% of the traces. Furthermore, we observed that performing joint inference over a person's state and location results in more consistent estimates. For instance, the person's location estimate is typically inside (or outside) a building whenever the environment state is estimated to be inside (or outside). This is not the case when GPS is not considered during inference.

## Virtual Evidence

All tests reported so far use fully labeled training data. In this section, we discuss some of our efforts on training us-ing virtual evidence (VE) for partially labeled data. Here, we focused on learning the parameters of activity recognition based on MSB information only (*i.e.*, no GPS information is used).

In order to test the robustness w.r.t. sparseness of data annotation, we proceeded as follows. The boosted classifiers were trained once based on a fixed set of training data. Such training can be performed without significant overhead, since it does not require any temporal structure or transitions between activities. When training the HMM, our experiment simulates a user who labels only a fraction of each activity in an activity sequence. To do so we first determined in the fully labeled data the time blocks in which an activity did not change. We then removed increasing amounts of training labels from the start and end of each activity block. Put differently, we removed increasing amounts of labels on either side of each transition between activities, as illustrated in Figure 6. This generated a sequence of fully labeled blocks that were interleaved with time periods that had no labels. Our experiment assumes that the user annotates at least one time frame in each activity in the sequence. Thus, between two labeled blocks, only the two activities of these blocks can occur. However, since the blocks have different durations, the transitions between activities are typically not in the middle between blocks.

The information about activities in the gaps between blocks was modeled using virtual evidence (see Section 3.3). To see this, let us assume that two consecutive blocks have activities $s^1$ and $s^2$. In the (unlabeled) region of uncertainty between them, we varied relative weights of $f(s_k = s^1)$ and $f(s_k = s^2)$ using different functions. These include a linear interpolation function, a uniform (equal) weight over the two activities on either side of the transition, and a uniform function over all the activities. Note that the uniform function over all activities corresponds to the standard semi-supervised setting in which some parts are fully labeled while the rest of the data has no labels at all.

Figure 7 shows the results of these experiments. The x-axis of this figure indicates the percent of frames dropped from each activity block (around each transition). While 0% corresponds to fully labeled data, 100% corresponds to a setting when only a single frame in the middle of each activity block is labeled. The thin, horizontal line gives the result obtained when using fully labeled data. The dotted line represents the result obtained when we made use of a linear interpolation function for VE, and the thick, solid line gives the result when using an equal value for the two activities between two labeled blocks. The dashed line indicates the results achieved when treating the gaps between labeled blocks as totally unlabeled, and training was performed using EM without any preference over the activities in these unlabeled regions.

As can be seen, when using VE the system performance is not effected to a large extent even when virtually all the labels were dropped! In other words, these results indicate that a user has to label only one frame for each activity he performs and still achieves virtually the same performance as a system trained with fully labeled data. This is a significant result as collecting fully labeled training data can be a highly challenging exercise. The dashed line in Figure 7 indicates that such a result cannot be achieved when using a standard approach in which the unlabeled data is trained without VE. In order to further test the above hypothesis, we repeated the experiment in which all but one label was dropped for each activity segment. This time, however, we chose the time point at which the label was set *randomly*, in contrast to choosing it based on the distance from the transitions. In other words, this simulates a practical system in which a user only labels one arbitrary frame for every activity. We ran this experiment 20 times and the mean accuracy was $81.69\%$ (with a $95\%$ confidence interval of $0.26$). Thus the system was able to achieve an accuracy which is not significantly below the $82.1\%$ achieved with the fully labeled training data!

In addition to making data annotation easier, VE is also able to achieve higher accuracy than the fully labeled data, as indicated by the increased accuracy when dropping 5% of the labels and using linear interpolation for VE (dotted line). This is due to the fact that the transition times between activities cannot always be labeled exactly. In such cases, VE is able to learn better transition times. The performance of VE with linear interpolation (dotted line) falls below the performance of VE with equal weight on two activities (thick solid line) as more data labels are dropped. This can be explained by the fact that the linear interpolation assumes that the transition between activities occurs in the middle of an unlabeled region, which may or may not be the case.

## 6 Conclusions

We presented an approach to estimate a person's low-level activities and spatial context using data collected by a small wearable sensor device. Our approach uses a dynamic Bayesian network to model the dependencies between the different parts of the system. This allows us to perform efficient inference and learning in the model. In experiments we show that the recognition accuracy of our system is significantly higher than the accuracy achieved by existing techniques that do not perform joint reasoning about a person's activities and spatial context. Furthermore, by taking information about the locations of buildings into account, our approach is able to correct GPS errors, thereby achieving more consistent estimates of a person's location and environment context.

We also show how virtual evidence (VE) can be used to model partially labeled data. The experiments indicate that

VE can yield better results than standard EM on fully labeled data. This is mostly due to the fact that VE can model uncertainty in the transition time between activities. The experiments additionally show that it is possible to train accurate HMM models by labeling only one time frame per activity in a training sequence.

While these results are extremely encouraging, they only present the first step toward fully recognizing a person's context. We believe that virtual evidence provides a powerful mechanism for smart data annotation. In future work, we intend to develop a modeling language and graphical user interface that uses VE to model various types of user annotations. For instance, we performed preliminary experiments in which VE is used to constrain the types of transitions occurring in a certain part of an activity sequence. The user can then specify, for example, that he ran at time $k$, drove a vehicle at time $k'$, and stopped exactly once for an unknown period of time in between.

Our experiments also showed that it is extremely difficult to handle outliers and biases in GPS, especially when a person moves between buildings. By improving the handling of GPS, we intend to make better use of the availability of maps and satellite images. Finally, our current system relies on a discretization of space in order to perform efficient inference and learning. We are also investigating the use of Rao-Blackwellised particle filters in order better represent a person's location in continuous space [18].

**Acknowledgments**

The authors would like to thank Tanzeem Choudhury, Jonathan Lester, and Gaetano Borriello for many useful discussions and for making their feature extraction and learning code available. Additional thanks go to the Intel Research Lab in Seattle for providing the sensor boards used in this research. This work has partly been supported by DARPA's ASSIST and CALO Programmes (contract numbers: NBCH-C-05-0137, SRI subcontract 27-000968), and by the NSF Human and Social Dynamics (HSD) program under contract number IIS-0433637.

## References

[1] D. Ashbrook and T. Starner. Using GPS to learn significant locations and predict movement across multiple users. *Personal and Ubiquitous Computing*, 7(5), 2003.

[2] L. Bao and S Intille. Activity recognition from user-annotated acceleration data. In *Proc. of the International Conference on Pervasive Computing and Communications*, 2004.

[3] J. Bilmes. On soft evidence in bayesian networks. Technical Report UWEETR-2004-0016, University of Washington, Dept. of EE, 2004.

[4] J. Bilmes and C. Bartels. On triangulating dynamic graphical models. In *Uncertainty in Artificial Intelligence: Proceedings of the Nineteenth Conference (UAI-2003)*, pages 47–56. Morgan Kaufmann Publishers, 2003.

[5] A. Blum and T. Mitchell. Combining labeled and unlabeled data with co-training. In *COLT: Proceedings of the Workshop on Computational Learning Theory*, 1998.

[6] O. Chapelle, A. Zien, and B. Schölkopf, editors. *Semi-supervised learning*. MIT Press, 2006.

[7] T. Dean and K. Kanazawa. Probabilistic temporal reasoning. In *Proc. of the National Conference on Artificial Intelligence (AAAI)*, 1988.

[8] R. Dechter. *Constraint Processing*. Morgan Kaufmann, 2003.

[9] J. Ho and S. Intille. Using context-aware computing to reduce the perceived burden of interruptions from mobile devices. In *Proc. of the Conference on Human Factors in Computing Systems (CHI)*, 2005.

[10] N. Kern, B. Schiele, H. Junker, P. Lukowicz, and G. Trster. Wearable sensing to annotate meeting recordings. In *In The 6th International Symposium on Wearable Computers (ISWC)*, 2002.

[11] N. Kern, B. Schiele, and A. Schmidt. Recognizing context for annotating a live life recording. *Personal and Ubiquituous Computingd*, 2005.

[12] J. Lester, T. Choudhury, N. Kern, G. Borriello, and B. Hannaford. A hybrid discriminative-generative approach for modeling human activities. In *Proc. of the International Joint Conference on Artificial Intelligence (IJCAI)*, 2005.

[13] L. Liao, D. Fox, and H. Kautz. Learning and inferring transportation routines. In *Proc. of the National Conference on Artificial Intelligence (AAAI)*, 2004.

[14] L. Liao, D. Fox, and H. Kautz. Location-based activity recognition. In *Advances in Neural Information Processing Systems (NIPS)*, 2005.

[15] P. Lukowicz, J. Ward, H. Junker, M. Stäger, G. Tröster, A. Atrash, and T. Starner. Recognizing workshop activity using body worn microphones and accelerometers. In *Proc. of Pervasive Computing*, 2004.

[16] D. Patterson, L. Liao, K. Gajos, M. Collier, N. Livic, K. Olson, S. Wang, D. Fox, and H. Kautz. Opportunity Knocks: a system to provide cognitive assistance with transportation services. In *International Conference on Ubiquitous Computing (UbiComp)*, 2004.

[17] J. Pearl. *Probabilistic Reasoning in Intelligent Systems: Networks of Plausible Inference*. Morgan Kaufmann Publishers, Inc., 1988.

[18] A. Raj, A. Subramanya, J. Bilmes, and D. Fox. Rao-Blackwellized particle filters for recognizing activities and spatial context from wearable sensors. In *Experimental Robotics: The 10th International Symposium*, Springer Tracts in Advanced Robotics (STAR). Springer Verlag, 2006.

[19] D. Yarowsky. Unsupervised word sense disambiguation rivaling supervised methods. In *Proceedings of the 33rd Annual Meeting of the Association for Computational Linguistics*, 1995.